\documentclass{Interspeech}

\interspeechcameraready

\title{Voice Adaptation for Swiss German}

\author{Samuel}{Stucki}
\author{Jan}{Deriu}
\author{Mark}{Cieliebak}

\affiliation[nocounter]{Centre for Artificial Intelligence}{ZHAW}{Switzerland}
\email{deri@zhaw.ch, stku@zhaw.ch, ciel@zhaw.ch}

\usepackage{comment}
\usepackage{url}
\usepackage{tabularx}
\begin{document}

\maketitle

\begin{abstract}
This work investigates the performance of Voice Adaptation models for Swiss German dialects, i.e., translating Standard German text to Swiss German dialect speech. For this, we preprocess a large dataset of Swiss podcasts, which we automatically transcribe and annotate with dialect classes, yielding approximately 5000 hours of weakly labeled training material. We fine-tune the XTTSv2 model on this dataset and show that it achieves good scores in human and automated evaluations and can correctly render the desired dialect. Our work shows a step towards adapting Voice Cloning technology to underrepresented languages. The resulting model achieves CMOS scores of up to -0.28 and SMOS scores of 3.8. 
\end{abstract}

\section{Introduction}
Voice Adaptation has seen an unprecedented increase in performance and utility in recent years~\cite{wang2023valle1,chen2024valle2}. It is now possible to clone a voice across languages with less than a minute of audio required~\cite{zhang2023vallex,casanova24xttsv2}. The two factors that led to these advancements are using large amounts of data and computing power, which are well-leveraged using the transformer architecture~\cite{vaswani2017transformers}. However, most of these advancements are made for high-resource languages such as English, Chinese, or German. The Vall-E models~\cite{wang2023valle1,chen2024valle2} were trained using over 50,000 hours of English audio data, and VallE-X was trained on an additional 10,000 hours of Chinese audio data. The data sources are high-quality readings of audio-books~\cite{kang2024libriheavy}. Recently, new datasets and pipelines, such as Emilia~\cite{he2024emiliaextensivemultilingualdiverse}, WenetSpeech4TTS~\cite{ma24d_interspeech}, and AutoPrep~\cite{yu2024autoprep} have further simplified the process of sourcing and preparing such data. The XTTS-v2 model~\cite{casanova24xttsv2} was trained on approx. 27,000 hours over 17 languages, with about half of the training material in English. Human evaluation was primarily done for English; the other languages were evaluated using automated methods. This shows a significant gap in the research between large and low-resource languages. 

In this work, we take the first step towards adapting voice adaptation technologies to the case of Swiss German, which has three main challenges. First, Swiss German refers to a group of dialects in the German-speaking part of Switzerland that can differ significantly in phonetics, vocabulary, and grammar. Second, these dialects are primarily spoken and lack a standardized written form. In written communication, speakers typically mix Standard German vocabulary with dialect-specific spelling and structures. Finally, Swiss German only has approximately five million speakers, distributed over a large variety of dialects, some of which are spoken by only several thousand speakers. 
For Speech-to-Text (STT), there has been considerable progress in transcribing and translating Swiss German speech into Standard German text. Most notably, the SDS200~\cite{pluss2022sds200} and the STT4SG-350~\cite{pluss2023stt4sg} corpora contain 200 and 350 hours of parallel data (i.e., Standard German text to Swiss German audio). These datasets propelled the performance for Speech-to-Text up to 77~\cite{timmel2024whisper4sg} BLEU score. On the other hand, the first systems for Swiss German Text-to-Speech (TTS) were enabled by the SwissDial corpus~\cite{dogan2021swissdial}. It covers eight dialects (with one speaker per dialect) and contains 2.5 to 4.55 hours of audio per dialect, totaling around 26 hours. They also created transcripts for each audio sample for each dialect. They trained a Tacatron 2 model~\cite{shen2018tacatron2} for each dialect separately and one joint model. Both models achieved MOS scores between 2.9 and 4.12, depending on the dialect. However, to our knowledge, no systematic research has been made to create a voice adaptation model for Swiss German so far. Such a Voice Adaptation solution for Swiss German would ideally take a Standard German text, a reference voice, and the desired dialect as input and generate speech in the desired dialect that renders the Standard German text. The main impediment is insufficient data currently available for Swiss German. Thus, the main investigations of this paper are guided by the following questions:
\begin{itemize}
\item Can high-quality STT for Swiss German be used to build an automated parallel corpus?
\item How well can a model trained on this data synthesize speech?
\end{itemize}

The main contribution of this paper is the answer to both of these questions. For this, we leverage the recent advances in speech technology to create an unsupervised parallel corpus and train a voice adaptation model based on that data. More concretely, we collect 5000 hours of automatically transcribed Swiss German speech with Standard German texts and fine-tune the XTTS-v2 model on this data. The resulting model achieves CMOS scores of up to $-0.28$ (almost human quality) and SMOS scores of 3.8 (very similar voices)~\footnote{We release the code upon acceptance.}. 
 
\section{Data Collection Pipeline}
\begin{figure}[t!]
  \centering
  \includegraphics[width=0.45\linewidth]{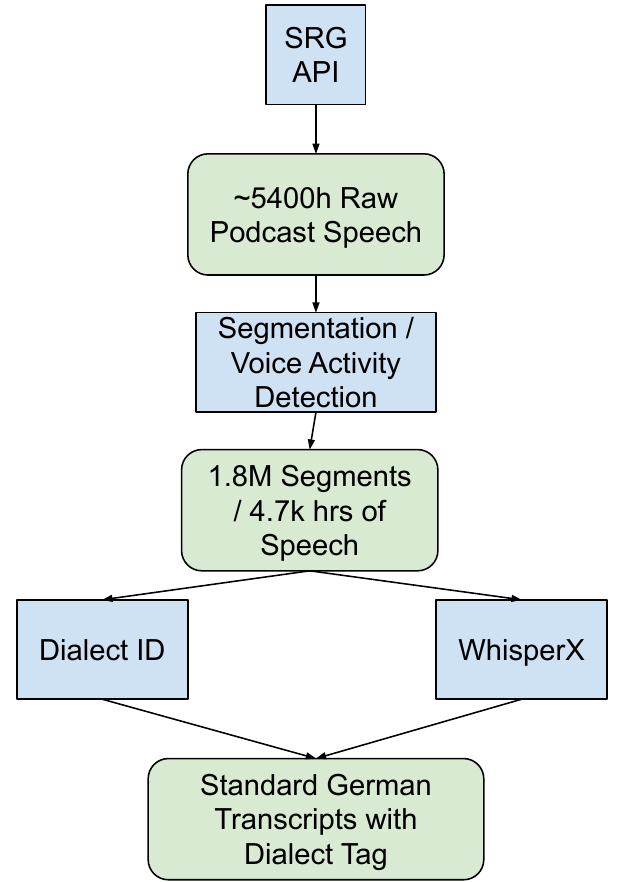}
  \caption{Data Processing Pipeline for pseudo labeled data.}
  \label{fig:data_pipe}
\end{figure}

In this section, we describe the creation of the pseudo-labeled dataset, which we ref to as SRG-corpus. We leverage data from the Swiss Broadcasting Cooperation (SRG) using their API~\footnote{\url{https://developer.srgssr.ch/}}, where we access available podcasts~\footnote{\url{https://www.srf.ch/audio/a-z}}. These podcasts cover various topics, from politics to entertainment and science. Figure~\ref{fig:data_pipe} shows the processing pipeline.

\subsection{Download}
We manually selected 25 podcasts out of 34, filtering out those programs that mainly contain music. Each podcast contains multiple episodes. Thus, each podcast's audio range varies between 9 hours and 1661 hours of raw audio. We manually classified the podcasts as being mainly Standard German speech (13), mainly Swiss German speech (6), or mixed (6) by listening to several episodes. We kept all podcasts, as the Standard German featured Swiss-accented speakers\footnote{Standard German spoken by Swiss Germans has a distinct accent recognizable by natives.}, and the larger dataset was expected to improve voice adaptation.
The dataset contains approximately 5,400 hours of raw audio. 

\subsection{Voice Activity Detection and Segmentation}
Next, we split each audio into segments with two requirements: each segment should only contain speech, and each segment should only contain speech from one speaker. Thus, we apply Voice Activity Detection (VAD) and Speaker Diarization to generate the segments. We used the pyannoteAI~\cite{bredin23diarization,plaquet23diarization} for both tasks.  Since the number of speakers per episode is unknown, we set a range of 2–6. While this may cause over-segmentation, it doesn't affect our pipeline, as we only require each segment to contain a single speaker.
Since the pipeline was mainly trained on English audio, we evaluated the Diarization Error Rate (DER) by manually annotating one episode of 42.63 minutes that contained three speakers using the ELAN~\cite{elan2024} software, which yielded a DER of 14.15\%, which corresponds to the DER values reported on the data pyannoteAI was trained on~\cite{plaquet23diarization}, and thus, is acceptable. 

We applied the Diarization Pipeline on the podcast episodes and filtered out any segment below 2 seconds. Since most TTS models work on segments of at most 15 seconds, we split any segment longer than 15 seconds into uniform 15-second segments. Note that this yields a disproportionate amount of samples of length equal to 15 seconds. However, Swiss German speech to Standard German text involves a translation part, making sentence-based segmentation difficult. This process yielded 1.81 million unique segments, each with a single speaker. 

\subsection{Transcription}
To create the transcripts, we used Whisper-Large-V3~\cite{radford2023whisper}, which has shown good performance in translating spontaneous Swiss German speech to Standard German text (achieving BLEU scores of 64.4) and can also handle examples with Standard German speech~\cite{timmel2024whisper4sg}. The transcription failed to produce a transcript for 2811 samples (0.15\%), and we removed these samples. To evaluate the quality of the transcripts, we manually transcribed 100 randomly sampled audio samples with Standard German texts. We computed the WER and BLEU scores between our annotations and the automatically generated transcripts. It yielded a median WER of 21.1\% and a BLEU score of 64.08, which lies in the expected range and is, thus, acceptable.  

\subsection{Dialect Identification}
\label{ssec:did}
To control the dialect during the generation, we automatically annotate the dialect of each sample. For this, we use the approach described in~\cite{bolliger2024automatische}, which is based on a pipeline that converts audio to phonemes using~\cite{xu2021wav2phoneme} and then trains a simple Naive Bayes model on top of phoneme n-grams. We use the same seven dialect regions as in~\cite{bolliger2024automatische}, which mirror the ones introduced in ~\cite{pluss2022sds200,pluss2023stt4sg}. We train the Naive Bayes n-gram model on the phonemized version of the STT4SG-350 corpus~\cite{pluss2023stt4sg}, which consists of approximately 30 hours of training data per dialect, and enrich the data with 30 hours of German Common Voice data~\cite{ardila2020commonvoice} (with equal gender distribution) to handle the Standard German parts of the SRG audio data. 
Following~\cite{bolliger2024automatische}, we concatenated samples of the same speaker in the test set to create 30s samples, achieving a macro F1-score of 0.88. The model often confuses the two geographically close regions, Zurich and Central Switzerland.  

We then applied this dialect identifier to all audio samples. For this, we merged the samples of the same speaker in an episode (using the diarization step from before) and then classified the speaker's dialect. 

\subsection{Corpus Overview}
Table~\ref{tab:srf_statistics} shows an overview of the data. The SRG data consists of 1.7M samples yielding 4,979 hours of audio. The distribution over the dialects is highly imbalanced: A third of the data consists of Standard German speech and a quarter consists of Zurich dialect data. Valais is the most underrepresented dialect, which is below 1\% of the total data (approx. 39 hours of speech). 

\begin{table}[t!]
\centering
\small
\resizebox{0.45\textwidth}{!}{%
\begin{tabular}{l|rrrr}
\hline
\textbf{Region}   & \textbf{Samples (K)} & \textbf{Length (h)} & \textbf{\% of Dataset} & \textbf{Tokens (M)} \\ \hline
Basel               & 179   & 460.81    & 9.25\%    & 5.35  \\ 
Bern                & 293	& 771.38    & 15.49\%   & 8.98  \\ 
German              & 538   & 1685.72   & 33.86\%   & 17.23  \\
Grisons             & 57	& 151.33    & 3.04\%   & 1.74   \\ 
Central CH          & 121	& 341.22    & 6.85\%   & 3.95    \\ 
Eastern CH          & 121	& 350.60    & 7.04\%   & 4.00    \\ 
Valais              & 15	& 39.46     & 0.79\%   & 0.43    \\ 
Zurich              & 440	& 1178.58   & 23.67\%  & 14.13   \\ \hline
Total               & 1,764	& 4,979.1   & 100\%    & 55.81  \\ \hline 
\end{tabular}}
\caption{SRG-corpus statistics by dialect concerning number of samples, duration, percentage of total duration, and number of tokens.}
\label{tab:srf_statistics}
\end{table}

\section{Model Training}
As a baseline, we fine-tuned the pre-trained XTTS-v2~\cite{casanova24xttsv2} on the STT4SG-350 data, which consists of 343 hours of training data, we refer to this model as~\emph{Baseline}. For our first model, we fine-tuned XTTS-v2 using the SRG and STT4SG-350 data mix. We refer to this as~\emph{SRG+STT4SG}. Since the SRG data is pseudo-labeled and contains noise due to spontaneous speech, while the STT4SG consists of highly controlled non-spontaneous speech samples, we create a third version where we further fine-tune  the~\emph{SRG+STT4SG} on the STT4SG-350 dataset; we refer to this as~\emph{SRG+STT4SG++}. Since the data was stored at 16kHz, we upsampled them to 22.5kHz~\footnote{Using pytorch audio https://pytorch.org/audio/stable/index.html}. The training was performed on 2 NVIDIA H200 GPUs with a per-device batch size of 36 and gradient accumulation steps of 14, resulting in a total batch size of 36 * 14 * 2 = 1008. We mostly applied the same setup as \cite{casanova24xttsv2} for the training setup, applying an AdamW optimizer with betas 0.9 and 0.96 and weight decay 0.01. The learning rate was changed to 6e-5 from the original 5e-5 due to internal tests and listening to the generated audio files. Weight decay was applied only to the weights, and the learning rate was decayed using MultiStepLR with a gamma of 0.5 using milestones 5000, 150000, and 300000. Table~\ref{tab:trained_models} offers an overview of the three models and the number of training steps.

\begin{table}[t!]
\centering
\small
\resizebox{0.45\textwidth}{!}{%
\renewcommand{\arraystretch}{0.95} 
\begin{tabularx}{\columnwidth}{lXc}
\hline
\textbf{Model} & \textbf{Description} & \textbf{Steps} \\ 
\hline
Baseline        & XTTSv2 fine-tuned on STT4SG.                & 170k  \\ 
SRG+STT4SG      & XTTSv2 fine-tuned on both STT4SG and SRG-corpus. & 238k  \\ 
SRG+STT4SG++    & SRG+STT4SG further fine-tuned on STT4SG.         & 170k  \\ 
\hline
\end{tabularx}}
\caption{Overview of trained models and their configurations.}
\label{tab:trained_models}
\end{table}

\section{Evaluation and Results}
We conducted two types of evaluations on the STT4SG-350 test set: automated and human. The training and test sets were pre-partitioned by the dataset authors to ensure speaker independence, such that no speaker or sample appears in both splits. The evaluation covers two scenarios.
\begin{itemize}
    \item \emph{Short}: Texts from the STT4SG-350 test set corresponding to 5-7 second utterances.
    \item \emph{Long}: Generated texts containing 2-4 sentences by ChatGPT-4o that correspond to utterances of around 10-15 seconds.
\end{itemize}
The decision to use generated texts in the Long scenario stems from the noisiness of the SRG corpus due to automated data collection.

\noindent\textbf{Automated Evaluation.} We evaluated three aspects:
\begin{itemize}
    \item Back-Translation Accuracy (i.e., translating the generated speech back to text): Measured using WER and BLEU scores between the text input to the TTS model and the back-translation by Whisper-Large-V3~\cite{radford2023whisper}.
    \item Speaker Similarity (SIM): Measured via cosine similarity using the ECAPA2 model~\cite{thienpondt2023ecapa2_ssim}.
    \item Dialect Recognition Accuracy (DID): Measured using our phoneme-based classification pipeline (c.f.~\ref{ssec:did}). 
\end{itemize}
We randomly sampled 50 Standard German texts for both scenarios, generating 1400 utterances (200 per dialect region). Five utterances for voice adaptation were randomly chosen for each speaker (four speakers per dialect).

\noindent\textbf{Human Evaluation.} We recruited six native speakers from Eastern CH (4), Central CH (1), and Bern (1) who evaluated the generated speech on the following metrics:
\begin{itemize}
    \item SMOS (Speaker Similarity): 1 (entirely dissimilar) to 5 (identical voice).
    \item CMOS (Comparative Mean Opinion Score): Comparison against reference speech on a scale from -3 (worse than the reference) to 3 (better than the reference).
    \item Intelligibility: Accuracy in rendering the target sentence from 1 (completely different) to 5 (exactly rendered the text).
\end{itemize}
Each scenario consisted of 42 utterances (6 per dialect), with two independent raters per sample (84 evaluations per scenario and system). Note that human evaluation of dialect accuracy is often unreliable due to varying human performance in dialect detection.

\begin{table}[t!]
\centering
\small
\resizebox{0.47\textwidth}{!}{%

\begin{tabular}{l|l|cccc}
\hline
\small
\textbf{Dialect} & \textbf{Model}   & \textbf{WER} & \textbf{BLEU}  & \textbf{SIM} & \textbf{DID}\\ \hline
            &   Baseline            & 0.424         & 0.497         &  0.456        & 0.981  \\
Basel       &   SRG+STT4SG          & 0.722         & 0.389         &  0.330        & 0.898  \\
            &   SRG+STT4SG ++       & 0.346         & 0.576         &  0.445        & 0.962  \\  \hline
            
            &   Baseline            & 0.372         & 0.550         &  0.459        & 1.000 \\
Bern        &   SRG+STT4SG          & 0.876         & 0.291         &  0.337        & 0.962  \\
            &   SRG+STT4SG ++       & 0.337         & 0.582         &  0.432        & 1.000 \\  \hline
            
            &   Baseline            & 0.285         &  0.635        &  0.446        & 0.962\\
Central CH  &   SRG+STT4SG          & 0.649         &  0.421        &  0.330        & 0.615  \\
            &   SRG+STT4SG ++       & 0.264         &  0.654        &  0.429        & 0.941\\  \hline
            
            &   Baseline            & 0.404         &  0.522        &  0.445        & 1.000\\
Eastern CH  &   SRG+STT4SG          & 0.809         &  0.336        &  0.327        & 1.000  \\
            &   SRG+STT4SG ++       & 0.366         &  0.557        &  0.439        & 1.000 \\  \hline
                
            &   Baseline            & 0.365         &  0.569        &  0.449        & 0.981\\
Grisons     &   SRG+STT4SG          & 0.765         &  0.353        &  0.331        & 0.875  \\
            &   SRG+STT4SG ++       & 0.377         &  0.557        &  0.428        & 1.000 \\  \hline
            
            &   Baseline            & 0.520         &  0.416        &  0.452        & 1.000\\
Valais      &   SRG+STT4SG          & 1.018         &  0.225        &  0.336        & 0.941  \\
            &   SRG+STT4SG ++       & 0.504         &  0.424        &  0.449        & 1.000 \\  \hline
            
            &   Baseline            & 0.371         &  0.558        &  0.445        & 0.313 \\
Zurich      &   SRG+STT4SG          & 0.760         &  0.353        &  0.332        & 0.412 \\
            &   SRG+STT4SG ++       & 0.326         &  0.599        &  0.429        & 0.457 \\  \hline \hline

            &   Baseline            & 0.391         &  0.533        &  0.441        &  -             \\
German      &   SRG+STT4SG          & 0.426         &  0.680        &  0.338        &  1.000        \\
            &   SRG+STT4SG ++       & 0.107         &  0.898        &  0.415        &  1.000        \\  \hline \hline

            &   Baseline            & 0.391         &  0.535        &  0.450        & 0.837          \\
Total       &   SRG+STT4SG          & 0.751         &  0.378        &  0.333        & 0.751     \\
            &   SRG+STT4SG ++       & 0.328         &  0.607        &  0.433        & 0.878     \\  \hline
\end{tabular}

}
\caption{Results of the automated evaluation in the \emph{Short} scenario.}
\label{tab:short_dialect_base}
\end{table}

\begin{table}[t!]
\centering
\small
\resizebox{0.47\textwidth}{!}{%

\begin{tabular}{l|l|cccc}
\hline
\small
\textbf{Dialect} & \textbf{Model}   & \textbf{WER} & \textbf{BLEU}  & \textbf{SIM} & \textbf{DID}\\ \hline
            &   Baseline            & 0.259         & 0.651         &  0.496   &  1.000\\
Basel       &   SRG+STT4SG          & 0.151         & 0.803         &  0.374   &  1.000  \\
            &   SRG+STT4SG ++       & 0.163         & 0.776         &  0.478   &  0.981\\  \hline
            
            &   Baseline            & 0.253         & 0.666         &  0.486   & 0.961\\
Bern        &   SRG+STT4SG          & 0.212         & 0.739         &  0.384   & 1.000   \\
            &   SRG+STT4SG ++       & 0.156         & 0.781         &  0.487   & 1.000 \\  \hline
            
            &   Baseline            & 0.204         &  0.726        & 0.482    & 1.000\\
Central CH  &   SRG+STT4SG          & 0.156         &  0.801        & 0.378    &  0.981  \\
            &   SRG+STT4SG ++       & 0.130         &  0.822        & 0.480    &  0.981\\  \hline
            
            &   Baseline            & 0.255         &  0.665        &  0.474   &  1.000\\
Eastern CH  &   SRG+STT4SG          & 0.174         &  0.780        &  0.378   &  1.000 \\
            &   SRG+STT4SG ++       & 0.170         &  0.763        &  0.482   &  1.000\\  \hline
                
            &   Baseline            & 0.240         &  0.685        &  0.475   &  1.000\\
Grisons     &   SRG+STT4SG          & 0.178         &  0.774        &  0.377   & 0.981  \\
            &   SRG+STT4SG ++       & 0.174         &  0.755        &  0.488   & 1.000 \\  \hline
            
            &   Baseline            & 0.325         &  0.583        &  0.485   &  1.000\\
Valais      &   SRG+STT4SG          & 0.316         &  0.634        &  0.380   &  0.897  \\
            &   SRG+STT4SG ++       & 0.284         &  0.613        &  0.493   &  1.000\\  \hline
            
            &   Baseline            & 0.238         &  0.688        &  0.482   &  0.200       \\
Zurich      &   SRG+STT4SG          & 0.182         &  0.770        &  0.377   &  0.138  \\
            &   SRG+STT4SG ++       & 0.144         &  0.798        &  0.477   &  0.071        \\  \hline \hline

            &   Baseline            & 0.196         &  0.727        & 0.477    &  -        \\
German      &   SRG+STT4SG          & 0.040         &  0.970        & 0.382    &  1.000   \\
            &   SRG+STT4SG ++       & 0.043         &  0.961        & 0.466    &  1.000         \\  \hline \hline

            &   Baseline            & 0.246         &  0.675        &  0.481   &  0.835     \\
Total       &   SRG+STT4SG          & 0.176         &  0.785        &  0.378   &  0.825 \\
            &   SRG+STT4SG ++       & 0.156         &  0.786        &  0.479   &  0.812  \\  \hline
\end{tabular}

}
\caption{Results of the automated evaluation in the \emph{Long} scenario.}
\label{tab:long_dialect_base}
\end{table}

\subsection{Automated Evaluation Results}
\noindent\textbf{Back Translation Results (WER \& BLEU).} We compared the back-translations to the input to the STT system and computed the WER~\footnote{\url{https://jitsi.github.io/jiwer/}}/BLEU (as per ~\cite{pluss2023stt4sg}). 
\begin{itemize}
    \item \emph{Short Results.} Table~\ref{tab:short_dialect_base} summarizes the results. The \emph{SRG+STT4SG++} model performs best (WER 0.328, BLEU 0.607), significantly outperforming the Baseline (WER 0.391, BLEU 0.535). The SRG+STT4SG model exhibits a much lower performance (WER 0.751, BLEU 0.378) due to artifacts introduced by the weakly labeled data. This has two causes: first, we trained mainly on Podcast data, which consists of many disfluencies of the speaker due to their spontaneous nature, which contrasts the single sentence nature of the \emph{Short} scenario, and second, that most samples in the SRG dataset are precisely 15 seconds of length, while the Short texts illicit speech of around 5-7 seconds.
    \item \emph{Long Results.} Table~\ref{tab:long_dialect_base} shows the results. Performance is better across all models, with \emph{SRG+STT4SG++} achieving WER 0.156 and BLEU 0.786. The \emph{SRG+STT4SG} model aligns more closely with \emph{SRG+STT4SG++} due to the long texts eliciting longer speech matching the SRG-corpus distribution.
\end{itemize}
Across dialects, Valais performs worst (WER 0.504, BLEU 0.424 in the \emph{Short} scenario) due to low data prevalence and is generally the most dissimilar dialect of all~\cite{paonessa2023dialect}. In contrast, Central CH performs best (WER 0.264, BLEU 0.654 in the \emph{Short} scenario), likely benefiting from Zurich’s high data prevalence. In general, Zurich and Bern exhibit similar performance, with Zurich scoring a WER of 0.144 and BLEU of 0.798 (\emph{SRG+STT4SG++} in the \emph{Long} scenario), while Bern scores a WER of 0.156 and BLEU of 0.781. Basel, Grisons, and Eastern CH models also perform comparably, showing strong BLEU scores above 0.750. 

\noindent\textbf{Speaker Similarity.} Using ECAPA2~\cite{thienpondt2023ecapa2_ssim}, we computed cosine similarity scores between reference and generated samples. The SIM column in Tables~\ref{tab:short_dialect_base} and~\ref{tab:long_dialect_base} shows the results. Scores range from 0.333 to 0.481 across both scenarios. That is, the speaker similarity is independent of the scenario. The \emph{Baseline} model performs best, probably due to the alignment with the STT4SG test set distribution. Conversely, the \emph{SRG+STT4SG} model performs worst, influenced by podcast prosody differences. Overall, all models achieve moderate speaker similarity, comparable to the scores reported in the  XTTSv2 report~\cite{casanova24xttsv2}.

\noindent\textbf{Dialect Recognition Accuracy.} Using our phoneme-based classification, dialect correctness was assessed on a speaker level by concatenating 50 generated utterances per speaker, as the classifier works well on utterances longer than 90 seconds. Thus, the evaluation is performed on a speaker level, not a sample level. The DID column in tables~\ref{tab:short_dialect_base} and~\ref{tab:long_dialect_base} shows the results. All models accurately render dialects with near-perfect classification. The exception is the Zurich dialect, which is underestimated, as the classifier often misassigns Zurich samples to Central Switzerland (c.f. Section~\ref{ssec:did}). However, in general, the dialects are correctly rendered.

\begin{table}[t!]
\centering
\resizebox{0.5\textwidth}{!}{%

\begin{tabular}{l| lcccc}
\toprule
\textbf{Eval Type} & \textbf{Model} & \textbf{SMOS} & \textbf{CMOS} & \textbf{Intelligibility} \\
\midrule
        & Baseline    &  3.10\(\pm\)$0.83$ & -0.80\(\pm\)$0.98$     &  4.17\(\pm\)$0.94$ \\
Short   & SRG+STT4SG  & \bf{3.39\(\pm\)$1.05^{*}$} & -0.55\(\pm\)$0.86$ &  3.85\(\pm\)$1.29^{\dagger}$  \\
        &SRG+STT4SG++ & 3.24\(\pm\)$0.90$ & \bf{-0.52\(\pm\)$0.84^{*}$} &  \bf{4.51\(\pm\)$0.69^{*\dagger}$} \\

\midrule
     & Baseline     &  2.98\(\pm\)$0.92$ & -0.94\(\pm\)$0.80$ &  3.96\(\pm\)$0.86$  \\
Long & SRG+STT4SG   &  \bf{3.81\(\pm\)$0.86^{*\dagger}$}      & \bf{-0.29\(\pm\)$0.74^{*\dagger}$} &   4.11\(\pm\)$0.89$  \\
     & SRG+STT4SG++ &  3.15\(\pm\)$0.92^{\dagger}$ & -0.69\(\pm\)$0.74^{*\dagger}$ & \bf{4.29\(\pm\)$0.70^{*}$}  \\
\bottomrule
\end{tabular}

}
\caption{Human evaluation result for the overall performance of the models in terms of mean $\pm$ standard deviation. * denotes that the SRG models are statistically different from the Baseline performance, and the $\dagger$ denotes that the two SRG model performances are statistically significantly different.}
\label{tab:human_eval}
\end{table}

\subsection{Human Evaluation Results}
Table~\ref{tab:human_eval} summarizes human evaluation results. 
\begin{itemize}
    \item \emph{Short Results.} \emph{SRG+STT4SG++} achieves the best CMOS and Intelligibility, confirming that this model better captures naturalness and accuracy in rendering speech. \emph{SRG+STT4SG} achieves the lowest intelligibility score, matching the automated evaluation, as it struggles with mismatches between its training data and the test set. However, it achieves the highest SMOS score, aligning with the automated speaker similarity results and indicating better voice consistency with reference speakers.
    \item \emph{Long Results.} The \emph{Baseline} performs worst across all categories, indicating that it does not generalize well to longer utterances. \emph{SRG+STT4SG} outperforms all models in SMOS and CMOS, suggesting that it provides the most speaker-consistent and natural speech synthesis in this setting. \emph{SRG+STT4SG++}  achieves the highest intelligibility score, indicating it produces the clearest and most accurate speech output.
\end{itemize}
Overall, the scores showcase that the human evaluation well receives the models trained on the SRG corpus. 

\subsection{Summary and Insights}
As confirmed by both automated and human evaluation, models trained on SRG data perform well overall in the \emph{Long} scenario. In the \emph{Short} scenario, \emph{SRG+STT4SG} struggles with the distribution mismatch of the SRG and STT4SG corpus. Overall, speaker similarity varies across models, but all retain moderate speaker identity. Finally, dialect correctness is generally high, except for Zurich misclassification due to classifier limitations. Thus, our model is a strong first step towards applying Voice Adaptation technologies to Swiss German dialects. 

\section{Conclusion}
This work presented the transfer of voice adaptation technologies to the challenging scenario of Swiss German dialects. We showed that translating Standard German text to Swiss German dialect speech is feasible and yields satisfactory results. For this, we created the SRG corpus of approximately 5000 hours of Swiss German dialect speech and fine-tuned an XTTSv2 model on this data. Finding a better approach for segmenting the speech to reflect sentences is expected to improve the performance even further.

\bibliographystyle{IEEEtran}
\bibliography{mybib}

\end{document}